\documentclass[10pt,twocolumn,letterpaper]{article}

\usepackage[pagenumbers]{iccv} % To force page numbers, e.g. for an arXiv version

\usepackage{amsmath}
\usepackage{algorithm}
\usepackage{listings}
\usepackage{times}

\usepackage{epsfig}
\usepackage{graphicx}
\usepackage{amssymb}
\usepackage{caption}
\usepackage{booktabs}
\usepackage{xcolor}
\usepackage{multirow}
\usepackage{colortbl}

\definecolor{darkgreen}{rgb}{0.0, 0.39, 0.0}
\definecolor{mygray}{gray}{0.92}
\definecolor{myred}{rgb}{0.8, 0.4, 0.4}

\definecolor{mygreen}{rgb}{0.65, 0.78, 0.54}
\definecolor{myorange}{rgb}{0.98, 0.67, 0.33}
\definecolor{myblue}{rgb}{0.46, 0.76, 0.87}
\definecolor{darkyellow}{rgb}{0.94, 0.71, 0.0}

\definecolor{iccvblue}{rgb}{0.21,0.49,0.74}
\usepackage[pagebackref,breaklinks,colorlinks,allcolors=iccvblue]{hyperref}

\title{Rethinking Bimanual Robotic Manipulation: \\ Learning with Decoupled Interaction Framework}

\author{
    \textbf{Jian-Jian Jiang}\textsuperscript{1}, 
    \quad \textbf{Xiao-Ming Wu}\textsuperscript{1},
    \quad \textbf{Yi-Xiang He}\textsuperscript{1}, 
    \quad \textbf{Ling-An Zeng}\textsuperscript{1}, \\
    \quad \textbf{Yi-Lin Wei}\textsuperscript{1},
    \quad \textbf{Dandan Zhang}\textsuperscript{5},
    \quad \textbf{Wei-Shi Zheng}\textsuperscript{1,2,3,4,}\footnotemark[2] \\
    \textsuperscript{1}  \small School of Computer Science and Engineering, Sun Yat-sen University, China \\
    \textsuperscript{2} \small Peng Cheng Laboratory, China \\
    \textsuperscript{3} \small Key Laboratory of Machine Intelligence and Advanced Computing, Ministry of Education, China \\
    \textsuperscript{4} \small Guangdong Province Key Laboratory of Information Security Technology, China \\
    \textsuperscript{5} \small Imperial-X Initiative and Department of Bioengineering, Imperial College London, U.K. \\
    {\tt\small \{jiangjj35, wuxm65\}@mail2.sysu.edu.cn; d.zhang17@imperial.ac.uk; wszheng@ieee.org}
}

\begin{document}
\maketitle

\begin{abstract}

    Bimanual robotic manipulation is an emerging and critical topic in the robotics community. Previous works primarily rely on integrated control models that take the perceptions and states of both arms as inputs to directly predict their actions. However, we think bimanual manipulation involves not only coordinated tasks but also various uncoordinated tasks that do not require explicit cooperation during execution, such as grasping objects with the closest hand, which integrated control frameworks ignore to consider due to their enforced cooperation in the early inputs. In this paper, we propose a novel decoupled interaction framework that considers the characteristics of different tasks in bimanual manipulation. The key insight of our framework is to assign an independent model to each arm to enhance the learning of uncoordinated tasks, while introducing a selective interaction module that adaptively learns weights from its own arm to improve the learning of coordinated tasks. Extensive experiments on seven tasks in the RoboTwin dataset demonstrate that: (1) Our framework achieves outstanding performance, with a \textbf{23.5\%} boost over the SOTA method. (2) Our framework is flexible and can be seamlessly integrated into existing methods. (3) Our framework can be effectively extended to multi-agent manipulation tasks, achieving a \textbf{28\%} boost over the integrated control SOTA. (4) The performance boost stems from the decoupled design itself, surpassing the SOTA by \textbf{16.5\%} in success rate with only \textbf{1/6} of the model size.    
    
\end{abstract}

\section{Introduction}
\label{sec:intro}

   Bimanual robotic manipulation \cite{bimanual_1, bimanual_3, household-services, medical-surgery, health-care, industrial-assembly}, which has strong capabilities to handle a wide range of complex tasks such as household services \cite{household-services}, medical surgery \cite{medical-surgery}, health care \cite{health-care} and industrial assembly \cite{industrial-assembly}, is an emerging and critical topic in the robotics community.

   \begin{figure}[t]
        \centering
        \includegraphics[width=0.9\linewidth]{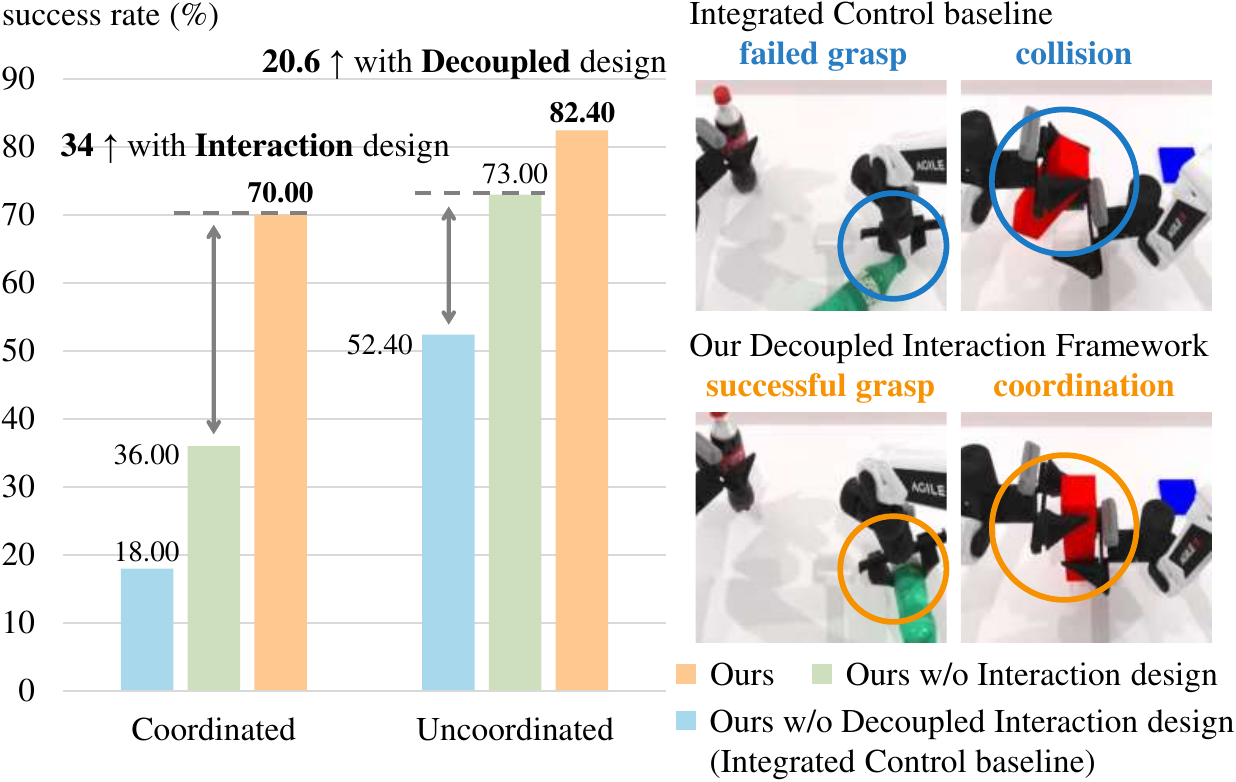}
        \caption{\textbf{Integrated Control vs. Decoupled Interaction.} The \textcolor{myblue}{blue} bar represents the success rate of coordinated and uncoordinated tasks for the integrated control baseline built upon our framework without decoupled interaction design. The \textcolor{mygreen}{green} and \textcolor{myorange}{orange} bars represent the success rate of coordinated and uncoordinated tasks for our framework without interaction design and our framework respectively. Our experiments are conducted on two coordinated tasks and five uncoordinated tasks in the RoboTwin dataset \cite{robotwin}. It can be observed that adding the decoupled design to the integrated control baseline promotes the learning of uncoordinated tasks. Furthermore, incorporating the interaction module on top of this design facilitates the learning of coordinated tasks.}
        \label{fig:motivation}
        \vspace{-5mm}
    \end{figure}

   Recently, with the help of imitation learning \cite{imitation_1, imitation_2}, bimanual robotic manipulation \cite{bimanual_1, bimanual_3, bimanual_4, bimanual_5, bimanual_6} achieves significant progress. Previous works typically use an integrated control model that takes observations and states of both arms as inputs and predicts actions for both arms simultaneously. However, based on taxonomy research on bimanual manipulation \cite{Taxonomy}, we think that bimanual robotic manipulation is more complex, involving not only coordinated tasks but also various uncoordinated tasks. In \textbf{uncoordinated tasks}, the two arms are neither spatially nor temporally coordinated and do not share directly connected goals. Each arm fulfills its task-specific constraints, with spatial coordination limited to avoiding collisions and no temporal coupling. For example, one arm holds a coffee cup while the other takes notes. In contrast, in \textbf{coordinated tasks}, both arms require spatial or temporal coordination, defined by specific task constraints. For instance, one arm places a block at a designated location, and the other positions another block on top afterward. Due to the unique characteristics of these tasks and the neglect of integrated control models to account for these differences, these models struggle to effectively learn uncoordinated and coordinated tasks. We conduct some confirmatory experiments using an integrated control baseline, as illustrated in Fig. \ref{fig:motivation}, which verifies our claim.
         
   \begin{figure}[t]
        \centering
        \includegraphics[width=0.9\linewidth]{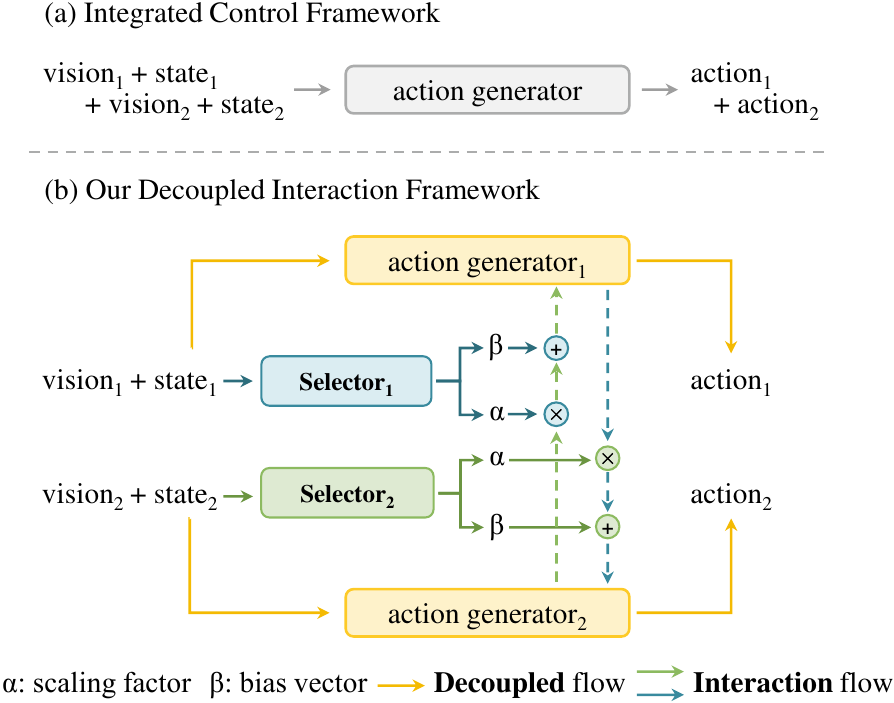}
        \caption{\textbf{Comparisons of our Decoupled Interaction Framework with integrated control frameworks.} Integrated control frameworks (a) mainly use a single model that takes the observations and states of both arms as inputs and directly outputs their actions. Our Decoupled Interaction Framework (b) first assigns an independent model to each arm to solely handle the inputs of the current arm (the \textcolor{darkyellow}{yellow} lines). Then, different from the naive interaction modeling in integrated control frameworks, a selective interaction module is proposed that learns its weights from its own arm to perform explicit modeling (the \textcolor{mygreen}{green} and \textcolor{myblue}{blue} lines) on the exchanged state features (the \textcolor{mygreen}{green} and \textcolor{myblue}{blue} dashed lines).}
        \label{fig:introduction}
        \vspace{-4mm}
    \end{figure}
   
   To address this, we propose learning bimanual manipulation via a precise task taxonomy \cite{Taxonomy} and introduce a novel Decoupled Interaction Framework for bimanual robotic manipulation. Compared to integrated control frameworks (as illustrated in Fig. \ref{fig:introduction} (a)), which directly predict actions for both arms, our framework (as illustrated in Fig. \ref{fig:introduction} (b)) first decouples the action dimensions the model needs to predict. Subsequently, unlike integrated control frameworks that primarily use the states of both arms for joint encoding to model the interaction between the two arms, our framework performs explicit interaction modeling between the two arms. Moreover, cooperation requirements vary at different stages, our framework selectively modulates interaction information during execution, enabling more effective use of this information. As illustrated in Fig. \ref{fig:motivation}, the integrated control baseline struggles to learn both uncoordinated and coordinated tasks, while incorporating the decoupled design and interaction design into the baseline effectively promotes learning of uncoordinated and coordinated tasks respectively. Notably, the decoupled design also facilitates learning of coordinated tasks, as it enhances the success rate of non-cooperative phases in coordinated tasks. Meanwhile, the interaction design also facilitates learning of uncoordinated tasks, as uncoordinated tasks still require minimal cooperation to avoid collisions between arms.

   Specifically, our framework first assigns an independent model to each arm and takes the observation and state of a single arm as inputs to generate actions only for that arm. This design is beneficial for uncoordinated tasks, where the action intentions and execution of each arm are relatively independent, enabling each arm to complete its own operations more effectively. Building upon this strong capability to handle uncoordinated tasks, our framework further employs explicit interaction modeling to improve cooperation, thereby facilitating the learning of coordinated tasks. To this end, our framework selects state features as interactive information and introduces a selective interaction module to adaptively modulate the exchanged features. First, our method uses a selective scaling module to adjust the intensity of the received state features. Then, to align the received state features with the current arm state space, a selective alignment module is introduced to adaptively generate a bias vector for the received state features. With our selective interaction module, our method satisfies the cooperation requirements of coordinated tasks.
  
   Extensive experiments on seven tasks in the RoboTwin dataset \cite{robotwin} demonstrate the following: (1) \textbf{Effectiveness}: Our framework achieves outstanding performance, obtaining a \textbf{23.5\%} improvement over the previous SOTA method \cite{single_2}. (2) \textbf{Flexibility}: Our framework can be seamlessly integrated into existing methods, such as DP3 \cite{single_2} and Point Flow Matching (PFM) \cite{generation1}. (3) \textbf{Extensibility}: Our framework can be effectively extended to multi-agent manipulation tasks, achieving a \textbf{28\%} improvement over the SOTA method \cite{single_2}. (4) \textbf{Scalability}: Experiments with an increased number of expert demonstrations confirm that our framework adheres to the scaling law. It should be emphasized that the performance improvement stems from the decoupled interaction design itself, surpassing the SOTA by \textbf{16.5\%} in performance with only \textbf{1/6} of the model size. We also conduct real-world experiments to further validate the effectiveness of our approach. Upon the publication of this paper, we will open-source the code.

\begin{figure}[t]
    \centering
    \includegraphics[width=0.95\linewidth]{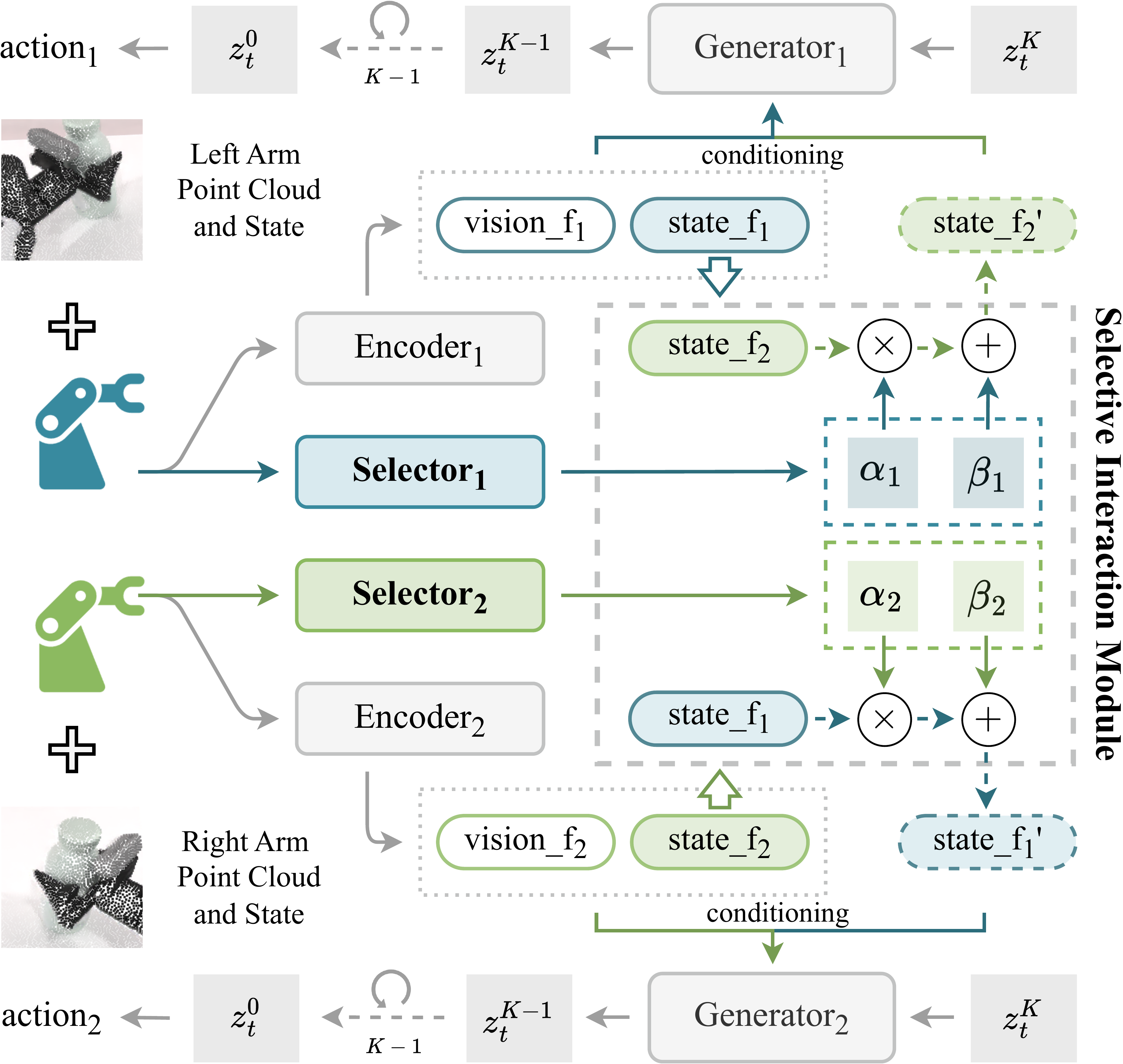}
    \caption{\textbf{Architecture of the Decoupled Interaction Framework.} Our framework first assigns a separate model to each arm to process its inputs. Then, we exchange state features between the models and utilize a selective interaction module to modulate them. Specifically, we use a selector to predict a scaling factor $ \alpha $ and a bias vector $ \beta $ to adaptively adjust the exchanged features. Finally, we combine the original visual features, state features and exchanged state features as interactive conditions to predict actions using action generators.}
    \label{fig:model}
    \vspace{-4mm}
\end{figure}

\section{Revisiting Bimanual Robotic Manipulation}

   Robotic manipulation, aiming to use extensive observation-action pairs from expert demonstrations to acquire human-like skills, is underpinned by grasp detection \cite{grasp_1, grasp_2, grasp_3, grasp_4, grasp_5, grasp_6}. Previous research mainly focuses on single-arm manipulation, where the modeling of policies is primarily divided into reinforcement learning \cite{reinforcement1, reinforcement2, reinforcement3, reinforcement4, reinforcement5, reinforcement6}, imitation learning \cite{regression1, regression2, regression3, regression4, regression5, regression6, single_1, single_2, single_3, generation1, generation2, generation3, generation4} and Vision-Language-Action (VLA) models \cite{VLA1, VLA2, VLA3, VLA4, VLA5}. Concretely, imitation learning methods mainly focus on regression-based \cite{regression1, regression2, regression3, regression4, regression5, regression6} and generation-based methods \cite{single_1, single_2, generation1, generation2, generation3, generation4}, which lay the foundation for policy designs of bimanual manipulation.

   Recently, bimanual manipulation becomes an emerging hotspot in the robotics community due to its ability to handle a wide range of tasks. First, we define this task. Given the observations $ O_{pcd} $ (point clouds as used in this paper) captured by depth cameras and the states of each arm $ S_{arm} $ provided by the ROS interface, bimanual manipulation aims to generate actions $ A_{arm} $ for the left and right arms to accomplish different categories of tasks.

   Benefiting from the development of imitation learning \cite{imitation_1, imitation_2}, bimanual manipulation \cite{bimanual_1, bimanual_2, bimanual_3, bimanual_4, bimanual_5, bimanual_6} methods achieve great progress. Previous works \cite{bimanual_1, bimanual_3, bimanual_4, bimanual_5, bimanual_6} mainly employ an integrated control framework that takes the observations and states of both arms as inputs and directly outputs their actions, i.e.:

  \vspace{-8pt}
  \begin{equation}
      \begin{split}
          A_{cent} &= \pi_{cent}(O_{pcd1}, O_{pcd2}, S_{arm1}, S_{arm2}), \\
          A_{cent} &\in \mathbb{R} ^ {1 \times 14}, O_{pcd} \in \mathbb{R} ^ {N \times 3}, S_{arm} \in \mathbb{R} ^ {1 \times 7}.
      \end{split}
  \end{equation}
  However, the high-dimensional task spaces hinder each arm from learning its own actions. This affects the efficiency for learning uncoordinated tasks, which focus more on whether each arm can complete its own actions. Moreover, the interaction modeling of integrated control frameworks is simple, where they typically perform joint encoding of the states of both arms. The implicit interaction modeling makes it difficult to learn coordinated tasks, where the cooperation requirements vary at different stages during execution.

  A few works like Voxact-b \cite{bimanual_2} utilize a fully decoupled framework for bimanual robotic manipulation. These works often take the observation and state of a single arm as inputs and output the actions of the current arm, i.e.:

  \vspace{-8pt}
  \begin{equation}
      \begin{split}
          &A_{arm1} = \pi_{dec}(O_{pcd1}, S_{arm1}), \\
          &A_{arm2} = \pi_{dec}(O_{pcd2}, S_{arm2}), \\
          &A_{arm} \in \mathbb{R} ^ {1 \times 7}, O_{pcd} \in \mathbb{R} ^ {N \times 3}, S_{arm} \in \mathbb{R} ^ {1 \times 7}.
      \end{split}
  \end{equation}
  These methods are usually applied to uncoordinated bimanual manipulation tasks because they do not explicitly model the collaboration between arms, making them less effective in handling coordinated tasks.

  In this paper, we propose to learn bimanual robotic manipulation through precise task taxonomy \cite{Taxonomy} and further introduce a Decoupled Interaction Framework. The insight of our framework is to utilize the decoupled design to reduce the action dimensions, thereby promoting the learning of uncoordinated tasks, while introducing a selective interaction module that adaptively modulates the interaction information to enhance the learning of coordinated tasks. Our experiments also comprehensively validate the excellent performance of our framework.

\section{A Decoupled Interaction Framework for Bimanual Robotic Manipulation}
In this section, we describe how our Decoupled Interaction Framework addresses the challenges in integrated control frameworks for bimanual manipulation. First, we revisit the motivation of our method, emphasizing that integrated control frameworks face issues of high-dimensional actions and a lack of explicit interaction modeling (Sec. \ref{subsec:motivation}). Next, we explain how our framework resolves these problems (Sec. \ref{subsec:decoupled} and Sec. \ref{subsec:interaction}). Finally, we summarize the overall structure of our framework (Sec. \ref{subsec:action}).

\subsection{Challenges in Integrated Control Frameworks}
\label{subsec:motivation}

The action intentions and execution of each arm are relatively independent in uncoordinated tasks. Therefore, it is important for methods to help each robotic arm effectively learn its own actions. However, most previous works do not adequately address this issue, since they mainly use an integrated control model to generate actions for both arms and combine the joint angle-based action spaces of each arm. This approach increases the dimensions of the actions the model needs to predict and enlarges the action space the neural network must explore, making the action learning processes for each arm more challenging.

Meanwhile, although cooperation between arms is important for coordinated tasks, this does not mean that both arms are constantly collaborating throughout the entire task. In other words, even in coordinated tasks, there are phases where each arm performs its own actions independently and phases where collaboration takes place. However, previous integrated control methods fail to effectively model the interaction between the arms, as they often rely on the states of both arms for joint encoding. This implicit modeling cannot distinguish different phases of coordinated tasks.

To validate it, we conduct experiments using our decoupled interaction framework. As shown in Fig. \ref{fig:motivation}, the integrated control baseline exhibits low performance, resulting in failed grasps and collisions between targets.

\subsection{The Proposed Decoupled Design Contributes to Uncoordinated Tasks}
\label{subsec:decoupled}

To address the problem of high-dimensional action spaces, in this paper, we propose to decouple the joint action space of both arms to reduce the space the neural network needs to explore, which helps our framework learn uncoordinated tasks more effectively. Specifically, our framework assigns an independent model to each arm, where it takes the observations and states of the current arm as inputs and outputs actions for the current arm. As shown in Fig. \ref{fig:motivation}, by incorporating the decoupled design, our framework outperforms the integrated control baseline in uncoordinated tasks. Moreover, as illustrated in Fig. \ref{fig:motivation}, our framework can effectively handle basic tasks like single-arm grasping, which is important in uncoordinated tasks. In conclusion, we consider that the decoupled design is beneficial for uncoordinated tasks. The loss of our decoupled design is:

\vspace{-12pt}
\begin{equation}
    \begin{split}
        \mathcal{L}_{total} &= \mathcal{L}_{arm1} + \mathcal{L}_{arm2}, \\
        \mathcal{L}_{arm1} &= \mathcal{L}_{action}(\theta_1, (O_{pcd1}, S_{arm1})), \\
        \mathcal{L}_{arm2} &= \mathcal{L}_{action}(\theta_2, (O_{pcd2}, S_{arm2})).
    \end{split}
\end{equation}
where $ \mathcal{L}_{action} $ represents the action generation loss, which will be described in Sec. \ref{subsec:action}. $ \theta_1 $ and $ \theta_2 $ represent the action generator for the left and right arm respectively.

It should be noted that although uncoordinated tasks require low cooperation, they still necessitate a certain level of interaction, i.e., the ability to perceive the state of the other arm to avoid collisions between the arms.

\vspace{-8pt}
\begin{equation}
    \begin{split}
        \mathcal{L}_{total} &= \mathcal{L}_{arm1} + \mathcal{L}_{arm2}, \\
        \mathcal{L}_{arm1} &= \mathcal{L}_{action}(\theta_1, (O_{pcd1}, S_{arm1}, F ^ {'}_{arm2})), \\
        \mathcal{L}_{arm2} &= \mathcal{L}_{action}(\theta_2, (O_{pcd2}, S_{arm2}, F ^ {'}_{arm1})).
    \end{split}
\end{equation}

\begin{table*}[t]
\resizebox{0.95\textwidth}{!}{
\begin{tabular}{c|cc|ccccc|c}
\hline
\multirow{2}{*}{Methods} & \multicolumn{2}{c|}{Coordinated Tasks} & \multicolumn{5}{c|}{Uncoordinated Tasks} & \multirow{2}{*}{\textbf{Average}} \\ \cline{2-8}
                          & \begin{tabular}[c]{@{}c@{}}Block\\ Handover\end{tabular} & \begin{tabular}[c]{@{}c@{}}Blocks\\ Stack\end{tabular} & \begin{tabular}[c]{@{}c@{}}Dual Bottles\\ Pick (Easy)\end{tabular} & \begin{tabular}[c]{@{}c@{}}Dual Bottles\\ Pick (Hard)\end{tabular} & \begin{tabular}[c]{@{}c@{}}Diverse\\ Bottles Pick\end{tabular} & \begin{tabular}[c]{@{}c@{}}Empty\\ Cup Place\end{tabular} & \begin{tabular}[c]{@{}c@{}}Block\\ Hammer Beat\end{tabular} & \\ \hline
Voxact-b \cite{bimanual_2} & 0.440 & 0.290 & 0.820 & 0.490 & 0.400 & 0.770 & 0.650 & \cellcolor{mygray} 0.551 (\textcolor{purple}{ $\downarrow$ 0.238}) \\ 
ACT \cite{bimanual_4}      & 0.070 & 0.020 & 0.340 & 0.040 & 0.020 & 0.310 & 0.310 & \cellcolor{mygray} 0.159 (\textcolor{purple}{ $\downarrow$ 0.630}) \\
DP \cite{single_1}         & 0.280 & 0.020 & 0.540 & 0.280 & 0.000 & 0.200 & 0.000 & \cellcolor{mygray} 0.189 (\textcolor{purple}{ $\downarrow$ 0.600}) \\
DP3 \cite{single_2}        & 0.700 & 0.230 & 0.780 & 0.460 & 0.380 & 0.730 & 0.600 & \cellcolor{mygray} 0.554 (\textcolor{purple}{ $\downarrow$ 0.235}) \\ \hline
Ours                       & \textbf{1.000} & \textbf{0.400} & \textbf{0.990} & \textbf{0.630} & \textbf{0.700} & \textbf{0.900} & \textbf{0.900} & \cellcolor{mygray} \textbf{0.789} \hspace{1.29cm} \\ \hline
\end{tabular}}
\caption{\textbf{Performance comparison of seven simulation tasks in the RoboTwin dataset.} Best results are highlighted in bold. Important comparison metrics are marked with gray cells. The red arrows indicate the performance difference between each baseline and our method.}
\label{table:robotwin}
\vspace{-8pt}
\end{table*}

\subsection{The Proposed Interaction Design Contributes to Coordinated Tasks}
\label{subsec:interaction}

To improve explicit interaction modeling, we build on the ability of our framework to handle uncoordinated tasks and introduce a novel selective interaction module. This module enhances interaction modeling and facilitates the learning of coordinated tasks. The key insight of our interaction module is to selectively modulate the state representations of the other arm by utilizing the observations and states of the current arm. This approach enables our framework to effectively control the influence of the other arm on each arm when each arm learns its own actions. As illustrated in Fig. \ref{fig:motivation}, by incorporating the selective interaction design, our framework outperforms the integrated control baseline in coordinated tasks. Moreover, as illustrated in Fig. \ref{fig:motivation}, our framework performs actions with better coordination.

\noindent \textbf{Scale the Intensity of Exchanged State Features.} Specifically, we exchange the encoded states between different arms and introduce a selective scaling module to scale the intensity of the received state representations. This module first processes the inputs of the current arm using modulation visual and state encoders. The encoded representations are then passed into an MLP, which predicts a scaling factor $ \alpha $ ranging from 0 to 1. Through this scaling factor $ \alpha $, our framework adaptively adjusts the influence of the state information from the other arm on the current arm across different phases of coordinated tasks.

\noindent \textbf{Align the Exchanged State Features.} Following that, to further align the received state representations with the state space of the current arm, we introduce a selective alignment module, which dynamically generates a bias vector for the received state representations. This module first takes the encoded observations and states from modulation visual and state encoders as inputs and then utilizes an MLP to predict a bias vector $ \beta $ for alignment.

To summarize, with this selective interaction module, the current arm can effectively filter and align the exchanged information from the other arm, thereby meeting the coordination requirements for both coordinated and uncoordinated tasks more effectively. The workflow of our selective interaction module is shown as follows:

\vspace{-8pt}
\begin{equation}
    \begin{split}
        \alpha_1, \beta_1 &= f_1(O_{pcd1}, S_{arm1}), g_1(O_{pcd1}, S_{arm1}), \\
        \alpha_2, \beta_2 &= f_2(O_{pcd2}, S_{arm2}), g_2(O_{pcd2}, S_{arm2}), \\
        F ^ {'}_{arm1} &= \alpha_2 \times F_{arm1} + \beta_2, \\
        F ^ {'}_{arm2} &= \alpha_1 \times F_{arm2} + \beta_1.
    \end{split}
\end{equation}
where $ F_{arm1} $ and $ F_{arm2} $ denote the exchanged state features, $ f(.) $ represents the regression head of scaling factors and $ g(.) $ represents the regression head of bias vectors. On the basis of the interaction design, the loss of our decoupled interaction framework can be updated as:

\subsection{A Decoupled Interaction Framework}
\label{subsec:action}
\noindent \textbf{Loss Design.} Our Decoupled Interaction Framework can be seamlessly integrated into existing methods, such as DP3 \cite{single_2} and Point Flow Matching \cite{generation1}. For DP3, we use DDIM \cite{ddim} as the action generator, and the ground truth is represented by the sample. For Point Flow Matching, we use Flow Matching \cite{cfm} as the action generator, and the ground truth is represented by the vector field. When employing DDIM, $ \mathcal{L}_{action} $ is represented as: 

\vspace{-10pt}
\begin{equation}
    \mathcal{L}_{action}(\theta, I) = \mathbb{E}_{t, z \sim \mathcal{D}(x)} \| z - \theta(z_t, t | I) \| ^ 2.
\end{equation}
where $ z \sim \mathcal{D}(x) $ denotes the distribution of the sample in demonstrations, $ I $ denotes the interactive condition and $ \theta $ denotes the DDIM action generator of each arm. When applying Flow Matching, $ \mathcal{L}_{action} $ is represented as: 

\vspace{-10pt}
\begin{equation}
    \begin{split}
        \mathcal{L}_{action}(\theta, I) &= \mathbb{E}_{t} \| \theta(x_t, t | I) - (x_1 - x_0) \| ^ 2, \\
        x_0 \sim p_0, x_1 &\sim p_1, x_t = tx_1 + (1 - t)x_0.
    \end{split}
\end{equation}
where $ p_0 $ represents a simple base density at time $ t = 0 $, $ p_1 $ represents the target complicated distribution at time $ t = 1 $, while $ x_0 $ and $ x_1 $ are the corresponding samplings. $ x_t $ is defined as the linear interpolation between $ x_0 $ and $ x_1 $, following the Optimal Transport theory \cite{COT}. $ \theta $ denotes the Flow Matching action generator of each arm, and $ I $ denotes the interactive condition.

\noindent \textbf{Overall Framework.} Bringing all together, our Decoupled Interaction Framework is developed. As shown in Fig. \ref{fig:model}, we first decouple the joint action space used in the integrated control framework by assigning an independent model to each arm, thereby promoting the learning of uncoordinated tasks. Then, we propose a brand new selective interaction module, which takes the inputs of the current arm to predict the modulation factors and modulates the exchanged state representations from the other arm, thereby promoting the learning of coordinated tasks. Finally, we combine the original observation and state representations with the modulated state representations as interactive conditions and utilize the action generator to predict actions for each arm. Benefiting from our decoupled interaction design, our framework effectively accommodates the unique characteristics of different tasks in bimanual manipulation, achieving significant improvements across various tasks.

\noindent \textbf{Implementation Details.} In this paper, we adopt the same point cloud backbone as DP3 \cite{single_2}. Following the setup in RoboTwin \cite{robotwin}, we utilize joint angles for proprioception and predict joint angles as actions. In the loss function, the weights for the left arm and right arm are both set to 1. Our framework is implemented using PyTorch and trained on a single NVIDIA RTX 4090 GPU for 3000 epochs with the AdamW optimizer and a batch size of 120. Additional implementation details are provided in Appendix \textcolor{myred}{5}.

\section{Experiments}
\subsection{Experiment Setups}
\noindent \textbf{Simulation Benchmark.} In this paper, we evaluate our framework on seven distinct manipulation tasks from the RoboTwin benchmark \cite{robotwin}. Based on the taxonomy of bimanual manipulation \cite{Taxonomy}, we define two tasks as coordinated tasks and five tasks as uncoordinated tasks. Each task is designed to assess specific aspects and detailed task descriptions can be found in Appendix \textcolor{myred}{6}.

\noindent \textbf{Coordinated Tasks.} (1) \textit{Block Handover}: Transferring the long block from the left arm to the right arm and placing it at the designated location. (2) \textit{Blocks Stack}: Stacking blocks of different colors in a specific order (the red block first and then the black block) to the designated location.

\noindent \textbf{Uncoordinated Tasks.} (1) \textit{Dual Bottles Pick (Easy)}: Lifting the bottles that are positioned randomly and standing upright simultaneously. (2) \textit{Dual Bottles Pick (Hard)}: Lifting the bottles that are positioned with random 6D poses simultaneously. (3) \textit{Diverse Bottles Pick}: Lifting the bottles that vary in random shapes and do not repeat in the training and testing sets simultaneously. (4) \textit{Block Hammer Beat}: Autonomously determining which arm picks up the hammer and strikes the block based on the position of the block. (5) \textit{Empty Cup Place}: Autonomously determining which arm picks up the cup and places it at the designated location based on the position of the cup.

\noindent \textbf{Baselines.} We select several representative approaches as our baselines. For regression-based integrated control approaches, we choose ACT \cite{bimanual_4} as a baseline, which takes multi-view images as inputs and directly regresses actions of both arms with an action chunking mechanism. For generation-based integrated control approaches, we choose and modify DP \cite{single_1}, which takes multi-view images as inputs and directly generates actions of both arms with receding horizon control \cite{RHC}, and DP3 \cite{single_2}, which takes 3D point clouds as inputs and directly generates actions like DP, as baselines. For decoupled approaches, we refer to the architecture of Voxact-b \cite{bimanual_2} and construct a decoupled baseline that takes point clouds as inputs and outputs the actions of each arm separately. More detailed implementations of all methods can be seen in Appendix \textcolor{myred}{5}.

\begin{table}[t]
\centering
\resizebox{\linewidth}{!}{
\begin{tabular}{c|c|c|c}
\hline
Decoupled & Scaling Factors & Bias Vectors & \textbf{Average} \\ \hline
                           &                           &                        & \cellcolor{mygray} 0.426  \hspace{1.29cm}                              \\
        \checkmark         &                           &                        & \cellcolor{mygray} 0.624 (\textcolor{darkgreen} {$\uparrow$ 0.198})   \\
        \checkmark         &       \checkmark          &                        & \cellcolor{mygray} 0.754 (\textcolor{darkgreen} {$\uparrow$ 0.328})   \\
        \checkmark         &                           &      \checkmark        & \cellcolor{mygray} 0.747 (\textcolor{darkgreen} {$\uparrow$ 0.321})   \\
        \checkmark         &       \checkmark          &      \checkmark        & \cellcolor{mygray} 0.789 (\textcolor{darkgreen} {$\uparrow$ 0.363})   \\ \hline
\end{tabular}}
\caption{\textbf{The ablation study.} Important metrics are in gray cells. The green arrows indicate the performance difference between each line and the first line (the baseline).}
\label{table:ablation}
\end{table}

\begin{table}[t]
\centering
\resizebox{\linewidth}{!}{
\begin{tabular}{c|ccc}
\hline
Task Name                & Demo-100 & Demo-150 & Demo-200 \\ \hline
Blocks Stack             & 0.470     & 0.530     & 0.650     \\
Dual Bottles Pick (Hard) & 0.660     & 0.740     & 0.850     \\ \hline
\end{tabular}}
\caption{\textbf{The scaling experiment on demonstration quantity and performance.} `` Demo-X '' indicates that X expert demonstrations are used during model training.}
\label{table:scaling}
\vspace{-8pt}
\end{table}

\subsection{Evaluation on RoboTwin Dataset}
\noindent \textbf{Quantitative Experiments.} We test our method and other representative methods on the RoboTwin \cite{robotwin} dataset and report the results in Tab. \ref{table:robotwin}. The models for different tasks are trained separately. From the table, we observe that: (1) Our method significantly outperforms previous methods. Compared to DP3 \cite{single_2}, the previous SOTA method, our method achieves an improvement of \textbf{23.5\%} in average success rate. (2) Specifically, our method surpasses DP3 by \textbf{23.5\%} in coordinated tasks and \textbf{23.5\%} in uncoordinated tasks respectively, which verifies the effectiveness of the decoupled interaction design in our framework. (3) In the ``Diverse Bottles Pick" task, the bottles used in the training and testing sets exhibit obvious differences in shape and texture. Achieving SOTA on this task verifies that our framework has great \textbf{intra-class generalization} abilities. (4) ``Block Hammer Beat'' and ``Empty Cup Place'' are single-arm manipulation tasks, where the challenge lies in the bimanual robot autonomously deciding which arm to use for manipulation based on the position of the object. Achieving a 90\% success rate on these tasks verifies that our model possesses great \textbf{decision-making} abilities.

\begin{figure*}[t]
    \centering
    \includegraphics[width=0.9\textwidth]{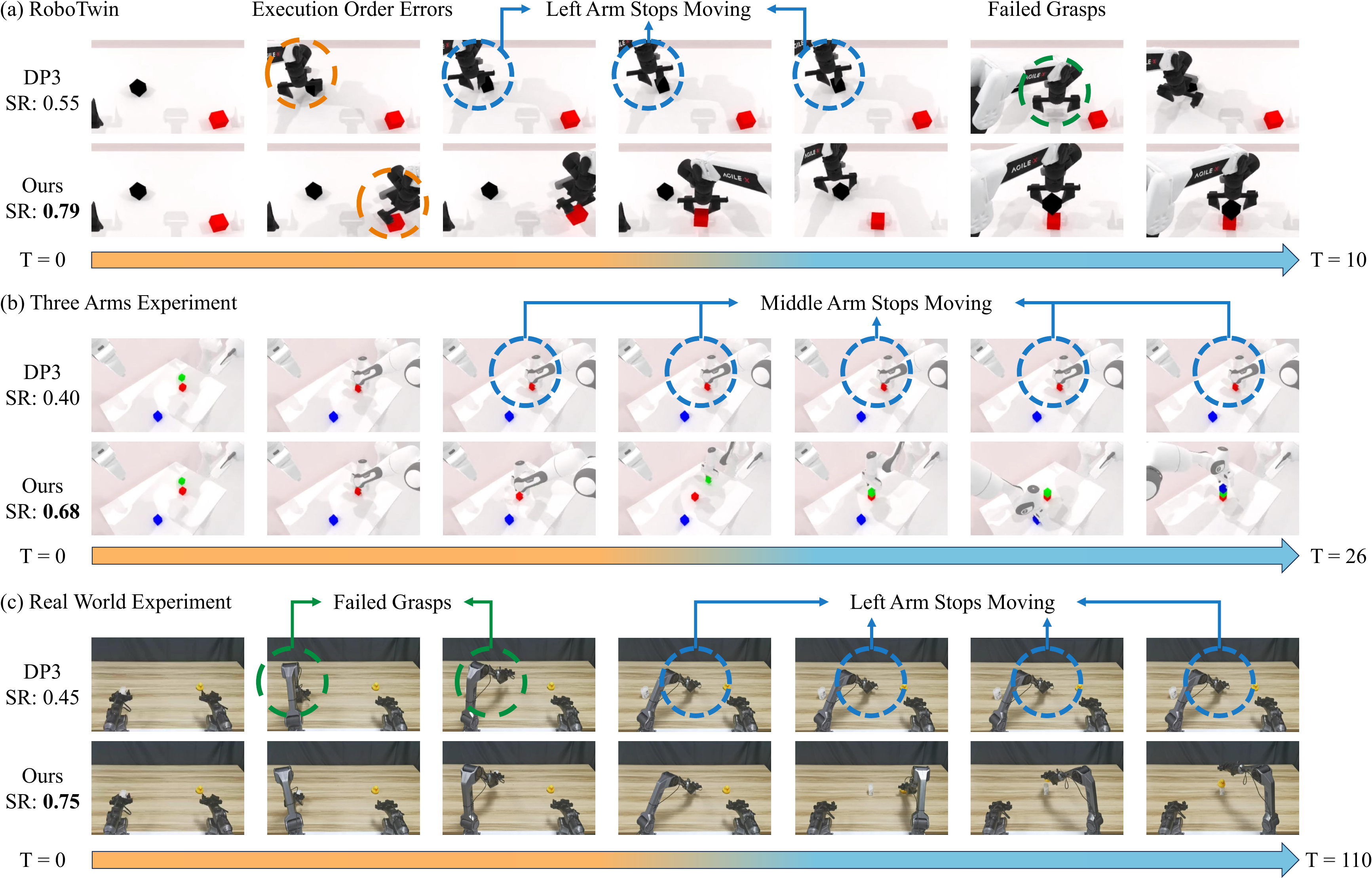}
    \caption{\textbf{Qualitative experiments.} In Fig. (a), we visualize the execution process of the ``Blocks Stack" task for DP3 and our framework. In Fig. (b), we visualize the execution process of the three-arm experiment for DP3 and our framework. In Fig. (c), we visualize the execution process of DP3 and our framework in the real-world experiment. Dashed circles of different colors highlight common issues that typically arise in integrated control frameworks. ``SR" denotes the success rate, which represents the average success rate of the model across all tasks under different experimental settings. Zoom in for a better view.}
    \label{fig:qualitative_experiments}
\vspace{-8pt}
\end{figure*}

\noindent \textbf{Qualitative Experiments.} Moreover, we also visualize the execution process of DP3 \cite{single_2} and our framework in the ``Blocks Stack" task. Both models are evaluated within the same scene configuration for consistency. As shown in Fig. \ref{fig:qualitative_experiments}, DP3 is more prone to failed grasps, while our model achieves more precise grasping. Additionally, for complex long-horizon tasks, DP3 often suffers from prolonged freezing behavior, where the arm remains stationary, as well as execution order errors. In contrast, our model can complete tasks in a more orderly manner. This is attributed to our decoupled interaction design, where the decoupled design decreases the action dimensions the model needs to predict, reducing the difficulty of network learning, and the interaction design modulates the exchanged state features of the other arm, effectively meeting the requirements for perceiving the other arm at different stages of coordinated tasks.

\begin{table}[t]
\centering
\resizebox{1\linewidth}{!}{
\begin{tabular}{c|cc|c}
\hline
Method                        & Coordinated        & Uncoordinated        & \textbf{Average}                                                                  \\ \hline
PFM \cite{generation1}        & 0.180              & 0.524                & \cellcolor{mygray} 0.426 \hspace{1.29cm}                                          \\
Ours (FM)                     & \textbf{0.700}     & \textbf{0.824}       & \cellcolor{mygray} \textbf{0.789} (\textcolor{darkgreen} {$\uparrow$ 0.363})      \\ \hline
DP3 \cite{single_2}           & 0.465              & 0.590                & \cellcolor{mygray} 0.554 \hspace{1.29cm}                                          \\
Ours (DDIM)                   & \underline{0.645}  & \underline{0.686}    & \cellcolor{mygray} \underline{0.674} (\textcolor{darkgreen} {$\uparrow$ 0.120})   \\ \hline
\end{tabular}}
\caption{\textbf{Integrating into existing methods experiments.} Best results are highlighted in bold. The underlined values indicate the second-best results. Important comparison metrics are marked with gray cells. ``FM'' denotes Flow Matching \cite{cfm}.}
\label{table:plug}
\end{table}

\begin{table}[t]
\centering
\resizebox{\linewidth}{!}{
\begin{tabular}{c|c|cc|c}
\hline
Method                & Model Size & Coordinated             & Uncoordinated             & \textbf{Average}                                \\ \hline
\multirow{2}{*}{DP3}  & 262.43M    & 0.465                   & 0.590                     & \cellcolor{mygray} 0.554                        \\
                      & 576.43M    & 0.150                   & 0.446                     & \cellcolor{mygray} 0.361                        \\ \hline
\multirow{3}{*}{Ours} & 42.95M     & 0.640                   & 0.750                     & \cellcolor{mygray} 0.719                        \\
                      & 146.37M    & \underline{0.665}       & \underline{0.788}         & \cellcolor{mygray} \underline{0.753}            \\
                      & 536.07M    & \textbf{0.700}          & \textbf{0.824}            & \cellcolor{mygray} \textbf{0.789}               \\ \hline
\end{tabular}}
\caption{\textbf{Model size and performance experiments.} Best results are highlighted in bold. The underlined values indicate the second-best results. Important metrics are marked with gray cells.}
\label{table:modelsize}
\vspace{-8pt}
\end{table}

\subsection{Ablation Study}
Furthermore, we conduct a series of ablation studies to verify the effectiveness of each design in our framework. All experiments are trained on seven tasks in the RoboTwin dataset \cite{robotwin}. In the ablation study, we choose Flow Matching \cite{cfm} as our action generator.

As illustrated in the second line of Tab. \ref{table:ablation}, the decoupled design enhances performance by 19.8\% in average success rate, demonstrating that the decoupled design is effective as it reduces the dimensions of actions and simplifies network learning. Moreover, as illustrated in the fifth line of Tab. \ref{table:ablation}, adding the interaction module, which includes scaling factors and bias vectors, to the decoupled design further improves performance by 16.5\% in average success rate. This demonstrates that the selective interaction design is beneficial, as it adaptively adjusts the exchanged information at different temporal stages during task execution. It should be noted that, from the third and the fourth lines of Tab. \ref{table:ablation}, although using scaling factors or bias vectors alone can effectively improve performance, they cannot achieve SOTA results. This is because they are complementary, i.e., only by scaling the intensity of the exchanged information and aligning it to the current state space together can the interaction information be fully modulated.

\subsection{Model Analysis}
\noindent \textbf{Adhering to the Scaling Law.} We analyze how the number of demonstrations affects the performance. We choose Flow Matching \cite{cfm} as the action generator and test different numbers of demonstrations used in the model training. The results are recorded in Tab. \ref{table:scaling}. From the table, it can be observed that as the number of demonstrations increases, the performance of our model continues to improve, indicating that our framework adheres to the scaling law.

\noindent \textbf{Integrating into Existing Methods Experiments.}
As mentioned in Sec. \ref{sec:intro}, our framework can be seamlessly integrated into existing methods. In this paper, we conduct experiments with two point cloud-based methods: DP3 \cite{single_2} and Point Flow Matching (PFM) \cite{generation1}. As illustrated in Tab. \ref{table:plug}, applying the proposed decoupled interaction design to integrated control frameworks using DDIM and Flow Matching significantly improves the success rate.

It should be noted that in DP3 \cite{single_2}, the performance of Flow Matching \cite{cfm} implemented with DPM Solver++ \cite{DPM} decreases compared to DDIM \cite{ddim}. The explanation provided in the DP3 paper is that the high dimensions of the task pose a challenge to the learning of DPM Solver++. For integrated control frameworks, as seen in the first and the third lines of Tab. \ref{table:plug}, the high dimensions of the tasks result in worse performance for PFM compared to DP3. From the second and fourth lines of Tab. \ref{table:plug}, it can be seen that our decoupled design effectively reduces the action space that Flow Matching needs to explore, thereby fully leveraging the generative abilities of Flow Matching.

\begin{figure}[t]
    \centering
    \includegraphics[width=0.9\linewidth]{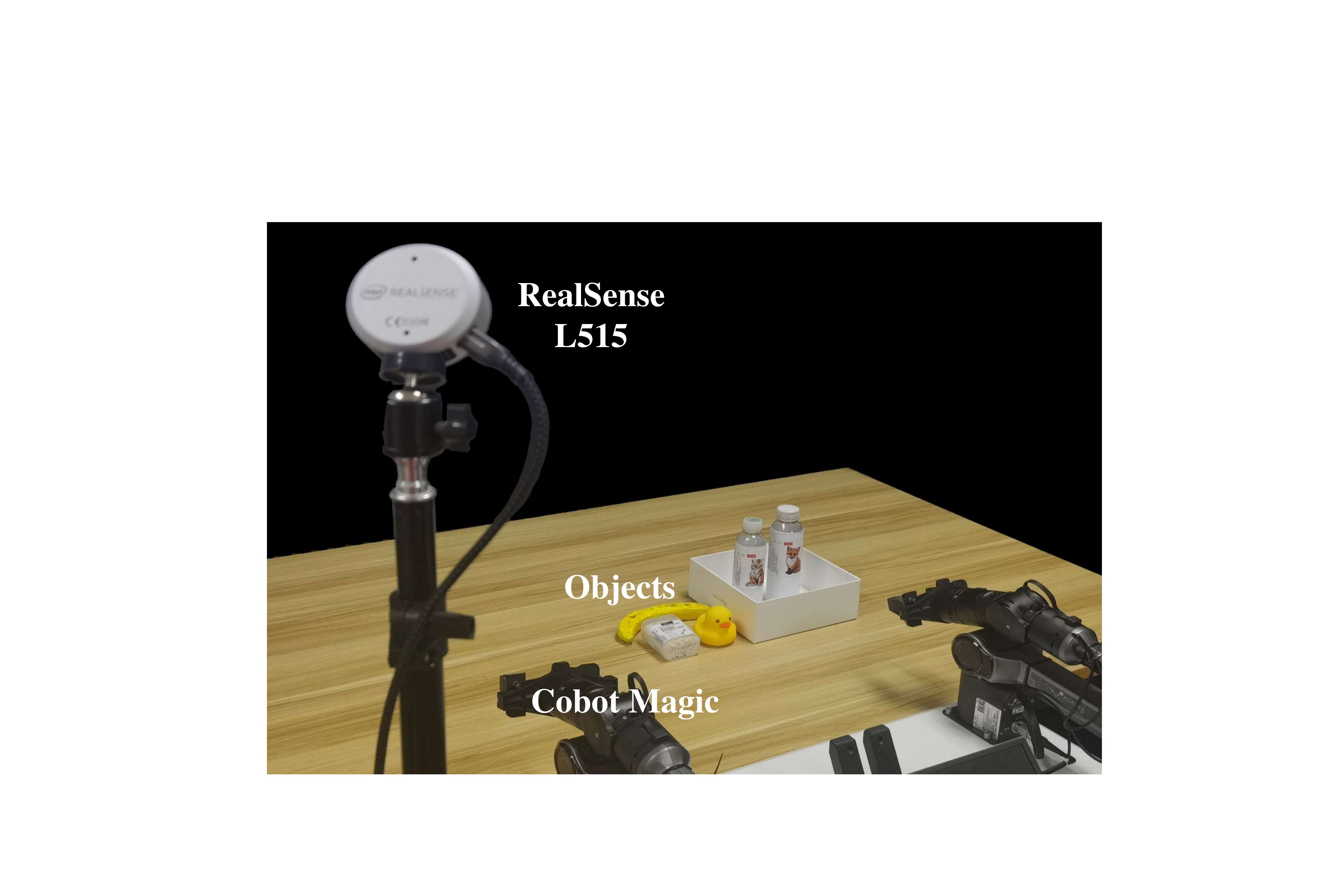}
    \caption{\textbf{Illustration of our real-world manipulation experimental settings.} We use the Cobot Magic as our bimanual robots and include everyday objects in our manipulation tasks. A RealSense L515 camera is applied to capture 3D point clouds.}
    \label{fig:setup}
\vspace{-8pt}
\end{figure}

\noindent \textbf{Model Size and Performance Experiments.}
We analyze the relationship between model size and performance, and record the results in Tab. \ref{table:modelsize}. Here, the action generator we adopt is Flow Matching \cite{cfm}. It can be concluded that: (1) From the first and second lines of Tab. \ref{table:modelsize}, it can be observed that when we increase the size of the action generator in DP3 \cite{single_2} to match the size of the action generator in our framework, the performance of DP3 significantly decreases by \textbf{19.3\%} compared to its original model size. (2) To test the performance of our framework with a smaller action generator, we compress the size of the action generator in our framework to match that of the action generator in DP3. As shown in the fourth and fifth lines of Tab. \ref{table:modelsize}, our model experiences only a 3.6\% performance drop despite a \textbf{73\%} reduction in the number of parameters. (3) As shown in the third line of Tab. \ref{table:modelsize}, when we further compress our model size to \textbf{1/6} of DP3 \cite{single_2}, our framework still outperforms DP3 by \textbf{16.5\%} in success rate. These experiments demonstrate that the primary reason for the improvement of our framework is not the increase in model parameters but rather the decoupled interaction design itself.

\noindent \textbf{Extensibility for Multi-Agent Manipulation.} Bimanual manipulation is the simplest form of multi-agent manipulation. To demonstrate the extensibility of our method, i.e., its applicability to broader tasks, we build a three-arm block stacking task based on the Sapien \cite{sapien} simulation. In this task, three robotic arms need to sequentially stack blocks of different colors in a specific order. In this experiment, a total of 50 demonstrations are utilized to train the two models independently. For evaluation, the success rate is measured for both models across 100 scenes.

As shown in Fig. \ref{fig:qualitative_experiments}, our framework achieves a \textbf{28\%} improvement compared to DP3. At the same time, the visualized trajectories indicate that as the dimensionality of the action space in integrated control frameworks further increases, the prolonged stationary state of the robotic arm after completing an action becomes more serious.

\subsection{Real-World Experiments.}
\noindent \textbf{Settings.} To verify the practical capabilities of our Decoupled Interaction Framework, we conduct real-world manipulation experiments across diverse tasks. For evaluation, we compare our framework with the integrated control baseline DP3 \cite{single_2} using 50 high-quality demonstrations collected via teleoperation similar to \cite{bimanual_4}.

\noindent \textbf{Coordinated Tasks.} (1) \textit{Banana Handover}: Picking the banana randomly placed on the left side of the table with the left arm, transferring it to the right arm, and placing it in the box.
(2) \textit{Items Stack}: Placing the box first and then the toy on top of the box in the designated location in order.

\noindent \textbf{Uncoordinated Tasks.} (1) \textit{Dual Bottles Pick}: Lifting the bottles that are positioned randomly and standing upright simultaneously. (2) \textit{Banana Pick and Place}: Autonomously determining which arm picks up the banana that is positioned randomly on the table and places it into the box.

\noindent \textbf{Quantitative Experiments.} The experiment setup and objects are shown in Fig. \ref{fig:setup}, and the testing results are shown in Tab. \ref{table:realworld}. It can be observed that our method achieves robust real-world manipulation ability, surpassing the SOTA method DP3 \cite{single_2} in real-world experiments, showing great potential for our Decoupled Interaction Framework.

\noindent \textbf{Qualitative Experiments.} We visualize the execution process of DP3 \cite{single_2} and our framework in ``Items Stack'' task. As shown in Fig. \ref{fig:qualitative_experiments}, DP3 often appears failed grasps, while our framework achieves more stable grasping. More examples and details can be seen in the video demo provided in the supplementary materials.

\begin{table}[t]
\centering
\resizebox{\linewidth}{!}{
\begin{tabular}{c|cc|cc|c}
\hline
\multirow{2}{*}{Method} & \multicolumn{2}{c|}{Coordinated} & \multicolumn{2}{c|}{Uncoordinated} & \multirow{2}{*}{\textbf{Total}} \\ \cline{2-5}
                        & \begin{tabular}[c]{@{}c@{}}Banana\\ Handover\end{tabular} & \begin{tabular}[c]{@{}c@{}}Items\\ Stack\end{tabular} & \begin{tabular}[c]{@{}c@{}}Dual \\ Bottels Pick\end{tabular} & \begin{tabular}[c]{@{}c@{}}Banana\\ Place\end{tabular} & \\ \hline
DP3                     &  6/15 &  5/15 &  6/15 & 10/15 &         27/60  \\
Ours                    & 10/15 & 10/15 & 13/15 & 12/15 & \textbf{45/60} \\ \hline
\end{tabular}}
\caption{\textbf{Real world experiments.} Best is in bold face.}
\vspace{-10pt}
\label{table:realworld}
\end{table}

\section{Conclusion}
In this paper, we propose learning bimanual manipulation through precise task categorization and introduce a novel Decoupled Interaction Framework for bimanual robotic manipulation. The key insight of our framework is to decouple the action space the network needs to explore by assigning an independent model to each arm, thereby enhancing the learning of uncoordinated tasks. Additionally, we propose a brand new selective interaction module that learns its weights from the inputs of its own arm to adaptively modulate the exchanged state features, thereby improving the learning of coordinated tasks. Extensive experiments on seven tasks in the RoboTwin dataset demonstrate that our framework exhibits \textbf{effectiveness}, \textbf{flexibility}, \textbf{extensibility}, and \textbf{scalability}. We believe our decoupled interaction framework provides insights for the community to further explore the modeling of bimanual and even multi-agent manipulation tasks. To advance the community, we will release our code upon the publication of the paper.

\vspace{0.3em}\noindent{\textbf{Acknowledgments.}} This work was supported partially by NSFC(92470202, U21A20471), Guangdong NSF Project (No. 2023B1515040025), Guangdong Key Research and Development Program(No.2024B0101040004).

{
    \small
    \bibliographystyle{ieeenat_fullname}
    \bibliography{main}
}

\clearpage
\begin{center}
\Large
Appendix
\end{center}
\setcounter{section}{0}

\section{Video Demo}
A video demo is provided for both simulation and real-world manipulation experiments using our Decoupled Interaction Framework, showing the \textbf{effectiveness}, \textbf{extensibility}, and \textbf{practical applicability} of our framework. In simulation experiments, we first collect 50 demonstrations for various tasks in the Sapien simulation environment. We then train our framework separately for each task. For real-world experiments, we utilize the Cobot Magic robotic arm and the RealSense L515 camera, employing single-view, third-person point clouds for model training and inference. Watch the video for more details. Enjoy and have Fun!

\begin{table*}[t]
\resizebox{0.95\textwidth}{!}{
\begin{tabular}{c|cc|ccccc|c}
\hline
\multirow{2}{*}{} 
    & \multicolumn{2}{c|}{Coordinated} 
    & \multicolumn{5}{c|}{Uncoordinated} 
    & \multirow{2}{*}{Average} \\ \cline{2-8}
    
    & \begin{tabular}[c]{@{}c@{}}Block\\ Handover\end{tabular} 
    & \begin{tabular}[c]{@{}c@{}}Blocks\\ Stack Easy\end{tabular} 
    & \begin{tabular}[c]{@{}c@{}}Dual Bottles\\ Pick Easy\end{tabular} 
    & \begin{tabular}[c]{@{}c@{}}Dual Bottles\\ Pick Hard\end{tabular} 
    & \begin{tabular}[c]{@{}c@{}}Diverse\\ Bottles Pick\end{tabular} 
    & \begin{tabular}[c]{@{}c@{}}Empty\\ Cup Place\end{tabular} 
    & \begin{tabular}[c]{@{}c@{}}Block\\ Hammer Beat\end{tabular} 
    &                          \\ \hline
    
MLP    & 1.000          & 0.330          & 0.960          & 0.550          & 0.590          & 0.860          & 0.860          & \cellcolor{mygray} 0.736 (\textcolor{purple}{$ \downarrow $ 0.053})         \\
Concat & 1.000          & 0.370          & 0.960          & 0.510          & 0.590          & 0.810          & 0.890          & \cellcolor{mygray} 0.733 (\textcolor{purple}{$ \downarrow $ 0.056})         \\
Ours   & \textbf{1.000} & \textbf{0.400} & \textbf{0.990} & \textbf{0.630} & \textbf{0.700} & \textbf{0.900} & \textbf{0.900} & \cellcolor{mygray} \textbf{0.789} \hspace{1.29cm}                           \\ \hline
\end{tabular}}
\caption{\textbf{Illustration of the performance with different interaction designs on seven tasks in the RoboTwin dataset.} Best results are highlighted in bold. Important comparison metrics are marked with gray cells. The red arrows indicate the performance difference between each baseline and our method.}
\label{table:selective}
\end{table*}

\section{Analysis of Selective Interaction Module}
We further explore various interaction design approaches. Specifically, we evaluate concatenation and MLP, which are widely adopted in the Computer Vision community for interaction modeling. The results are presented in Tab. \ref{table:selective}. As shown in Tab. \ref{table:selective}, compared to the MLP and concatenation-based interaction modeling, our model achieves a performance improvement of at least 5.3\%, demonstrating the effectiveness of our selective interaction module.

\section{Visualization of the Generated Manipulation Trajectories in RoboTwin}
\begin{figure*}[t]
    \centering
    \includegraphics[width=\textwidth]{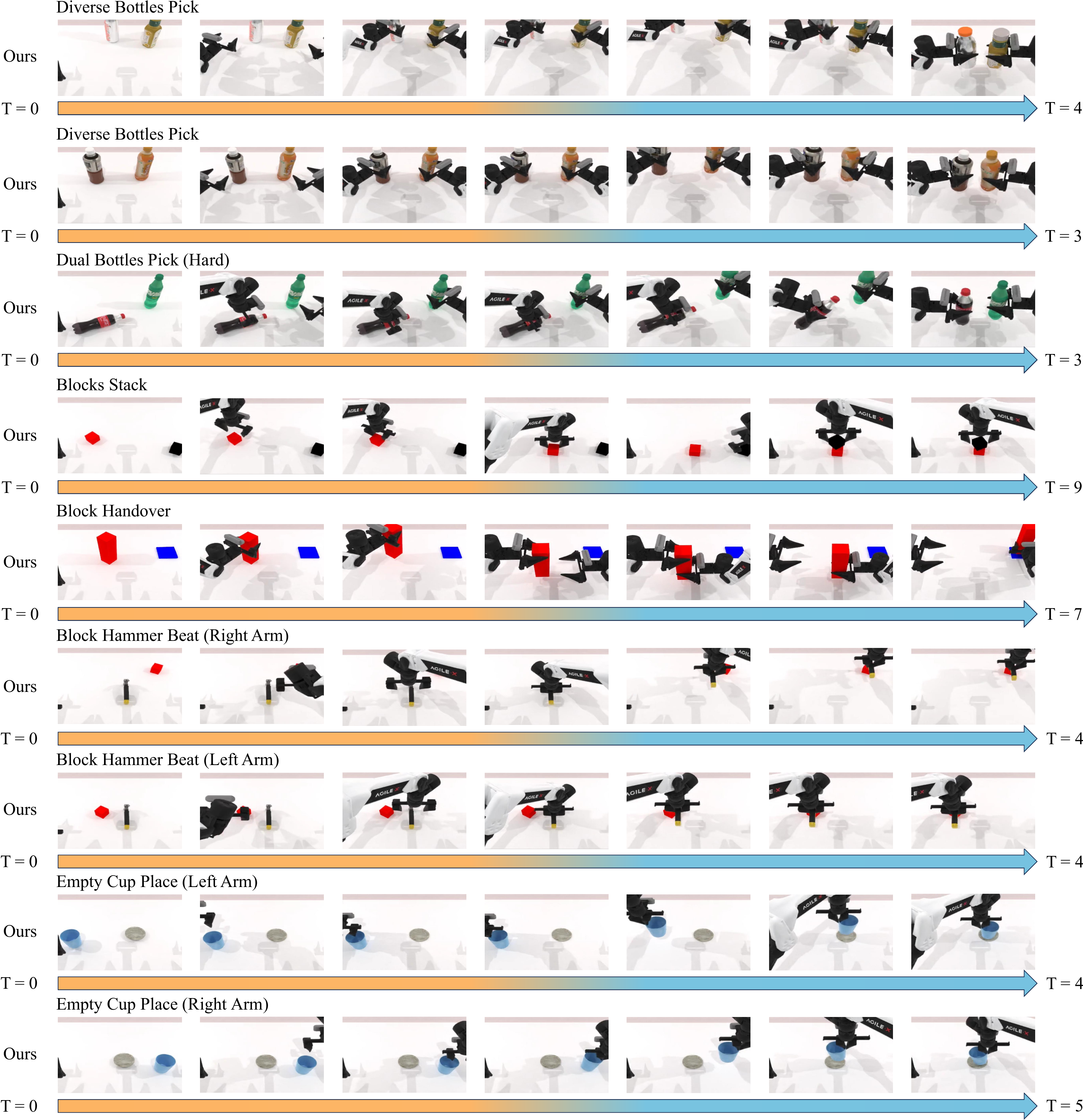}
    \caption{\textbf{Visualization of the generated manipulation trajectories of our framework in RoboTwin.} We visualize different coordinated and uncoordinated tasks within various scenes. Zoom in for best view.}
    \label{fig:supp_qualitative}
\end{figure*}

In this section, we visualize some manipulation trajectories in RoboTwin \cite{robotwin} generated by our Decoupled Interaction Framework. As illustrated in Fig. \ref{fig:supp_qualitative}, it can be concluded that: (1) We visualize two manipulation trajectories with different target objects in the ``Diverse Bottles Pick" task to demonstrate the capability of our framework for \textbf{intra-class generalization}. (2) Additionally, we visualize trajectories for the ``Block Hammer Beat" and ``Empty Cup Place" tasks with varying initial positions of the target objects. As shown in Fig. \ref{fig:supp_qualitative}, our model autonomously decides which arm to use based on the position of the target object, demonstrating its \textbf{decision-making} ability.

\section{Visualization of the Generated Manipulation Trajectories in Real World}
\begin{figure*}[t]
    \centering
    \includegraphics[width=\textwidth]{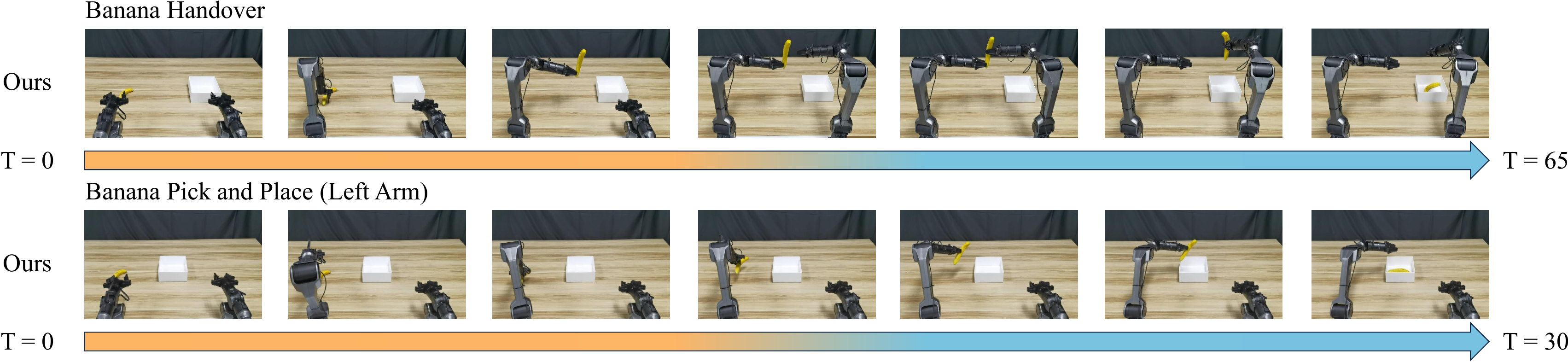}
    \caption{\textbf{Visualization of the generated manipulation trajectories of our framework in real-world experiments.} We visualize coordinated and uncoordinated tasks under different configurations. Zoom in for the best view.}
    \label{fig:supp_real}
\end{figure*}

In this section, we visualize some manipulation trajectories in real world generated by our Decoupled Interaction Framework. As illustrated in \ref{fig:supp_real}, our framework effectively handles both coordinated and uncoordinated tasks. This is because our decoupled design effectively reduces the high-dimensional action space, enabling the network to learn actions more efficiently. Additionally, our selective interaction module explicitly models the interaction between the arms, allowing our framework to better meet the cooperation requirements of various bimanual manipulation tasks.

\section{Implementation Details}
In this section, we provide a detailed introduction to the implementation details of all baselines and our Decoupled Interaction Framework.

\noindent \textbf{Training Setup.} The key training setup for our Decoupled Interaction Framework based on the DDIM \cite{ddim} and Flow Matching \cite{cfm} is detailed in Tab. \ref{tab:ours_setup}.

\noindent \textbf{Baseline Setups.} We also outline the training settings for the baseline in Tab. \ref{tab:baseline_setup}. Because of the differences in hyper-parameters between ACT and other baselines, we provide a description of its hyper-parameters here. For the ACT method, we use the AdamW optimizer with an initial learning rate of 1.0e-5 and a weight decay of 1.0e-4. The training process employs a batch size of 8, runs for 2000 epochs and uses an action chunking size of 100.

\section{Simulation Tasks}

\begin{figure*}[t]
    \centering
    \includegraphics[width=\textwidth]{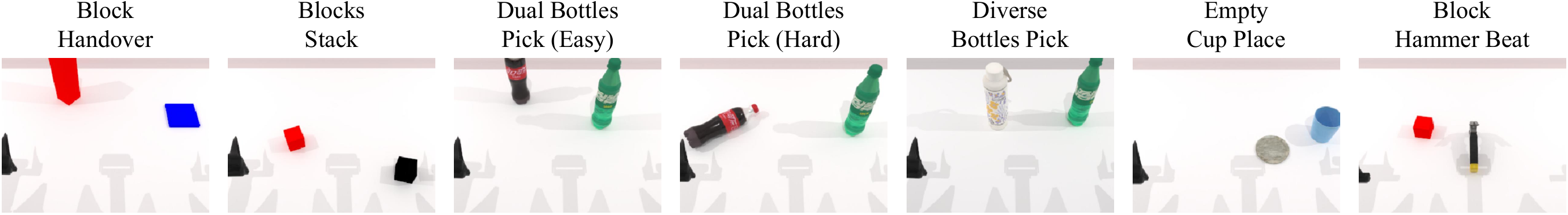}
    \caption{\textbf{Seven testing benchmark tasks.} We visualize manipulation tasks used in the RoboTwin benchmark. Zoom in for the best view.}
    \label{fig:task_visual}
\end{figure*}

We also visualize the distinct manipulation tasks from the RoboTwin benchmark \cite{robotwin}, as illustrated in Fig. \ref{fig:task_visual}, and provide detailed descriptions of all simulation tasks in Tab. \ref{tab:benchmark_description}, totaling seven tasks.

\begin{table*}[t]
\small
\centering
\begin{tabular}{ccc}
\midrule
Parameter & Ours (DDIM\cite{ddim}) & Ours (Flow Matching\cite{cfm}) \\ 
\midrule
horizon & 8 & 8 \\ 
n\_obs\_steps & 3 & 3 \\ 
n\_action\_steps & 6 & 6 \\ 
num\_inference\_steps & 10 & 10 \\ 
dataloader.batch\_size & 120 & 120 \\ 
dataloader.num\_workers & 8 & 8 \\ 
dataloader.shuffle & True & True \\ 
dataloader.pin\_memory & True & True \\ 
dataloader.persistent\_workers & False & False \\ 
optimizer.\_target\_ & torch.optim.AdamW & torch.optim.AdamW \\ 
optimizer.lr & 1.0e-4 & 3.0e-5 \\ 
optimizer.betas & [0.95, 0.999] & [0.95, 0.999] \\ 
optimizer.eps & 1.0e-8 & 1.0e-8 \\ 
optimizer.weight\_decay & 1.0e-6 & 1.0e-6 \\ 
training.lr\_scheduler & cosine & cosine \\ 
training.lr\_warmup\_steps & 500 & 10 \\ 
training.num\_epochs & 3000 & 3000 \\ 
training.gradient\_accumulate\_every & 1 & 1 \\ 
training.use\_ema & True & True \\ 
\bottomrule
\end{tabular}
\vspace{-3pt}
\caption{\textbf{Model training settings.} Hyper-parameter Settings for Training and Deployment of our Decoupled Interaction Framework.}
\label{tab:ours_setup}
\end{table*}

\begin{table*}[t]
\small
\centering
\begin{tabular}{cccc}
\midrule
Parameter & DP \cite{single_1} & DP3 \cite{single_2} & Voxact-b \cite{bimanual_2} \\ 
\midrule
horizon & 8 & 8 & 8 \\ 
n\_obs\_steps & 3 & 3 & 3 \\ 
n\_action\_steps & 6 & 6 & 6 \\ 
num\_inference\_steps & 100 & 10 & 10 \\ 
dataloader.batch\_size & 128 & 256 & 256 \\ 
dataloader.num\_workers & 0 & 8 & 8 \\ 
dataloader.shuffle & True & True & True \\ 
dataloader.pin\_memory & True & True & True \\ 
dataloader.persistent\_workers & False & False & False \\ 
optimizer.\_target\_ & torch.optim.AdamW & torch.optim.AdamW & torch.optim.AdamW \\ 
optimizer.lr & 1.0e-4 & 1.0e-4 & 1.0e-4 \\ 
optimizer.betas & [0.95, 0.999] & [0.95, 0.999] & [0.95, 0.999] \\ 
optimizer.eps & 1.0e-8 & 1.0e-8 & 1.0e-8 \\ 
optimizer.weight\_decay & 1.0e-6 & 1.0e-6 & 1.0e-6 \\ 
training.lr\_scheduler & cosine & cosine & cosine \\ 
training.lr\_warmup\_steps & 500 & 500 & 500 \\ 
training.num\_epochs & 300 & 3000 & 3000 \\ 
training.gradient\_accumulate\_every & 1 & 1 & 1 \\ 
training.use\_ema & True & True & True \\ 
\bottomrule
\end{tabular}
\vspace{-3pt}
\caption{\textbf{Baselines settings.} Hyper-parameter Settings for Training and Deployment of DP, DP3 and Voxact-b Algorithms.}
\label{tab:baseline_setup}
\end{table*}

\begin{table*}[t]
\centering
\small

\begin{tabular}{>{\itshape\centering}m{3.5cm} p{8cm}} 
\toprule
\textbf{Task}   & \textbf{Description}        \\ \midrule 
Block Handover & A long block is placed on the left side of the table. The left arm grasps the upper side of the block and then hands it over to the right arm, which places the block on the blue mat on the right side of the table. \\ \midrule
Blocks Stack Easy & Red and black cubes are placed randomly on the table. The robotic arm stacks the cubes in order, placing the red cubes first, followed by the black cubes, in the designated target location. \\ \midrule
Dual Bottles Pick Easy & A red bottle is placed randomly on the left side, and a green bottle is placed randomly on the right side of the table. Both bottles are standing upright. The left and right arms are used simultaneously to lift the two bottles to a designated location. \\ 
\midrule
Dual Bottles Pick Hard & A red bottle is placed randomly on the left side, and a green bottle is placed randomly on the right side of the table. The bottles' postures are random. Both left and right arms are used simultaneously to lift the two bottles to a designated location. \\ \midrule
Diverse Bottles Pick & A random bottle is placed on the left and right sides of the table. The bottles' designs are random and do not repeat in the training and testing sets. Both left and right arms are used to lift the two bottles to a designated location. \\ \midrule
Empty Cup Place & An empty cup and a cup mat are placed randomly on the left or right side of the table. The robotic arm places the empty cup on the cup mat. \\ \midrule
Block Hammer Beat & There is a hammer and a block in the middle of the table. If the block is closer to the left robotic arm, it uses the left arm to pick up the hammer and strike the block; otherwise, it does the opposite. \\  \bottomrule

\end{tabular}
\caption{\textbf{Task descriptions for RoboTwin platform.}}
\label{tab:benchmark_description}
\end{table*}

\end{document}